\documentclass{mcmthesis}  % 文档类型

\mcmsetup{CTeX = true,   % 使用 CTeX 套装时，设置为 true
        tcn =  2524697,   % 队伍控制号
        problem = C,  % 选题
        sheet = false,   % sheet页
        titleinsheet = false,   % sheet页显示标题
        keywordsinsheet = false,  % sheet页显示关键词
        titlepage = true,   % 标题页
        abstract = true  % 摘要
        }
\usepackage{palatino}  % 帕拉提诺体字体宏包
\usepackage{lipsum}  % 导入生成段落的宏包
\usepackage[hyperref=true,style=ieee]{biblatex}  % biblatex参考文献宏包
\usepackage{booktabs}
\usepackage{setspace}
\usepackage{titlesec}
\usepackage{authblk}  % 添加这个包
\vspace{-15pt}
\title{Exploring Patterns Behind Sports}  % 文章标题
\vspace{20pt}  % 增加标题和作者之间的距离
\author{\normalsize Chang Liu, Chengcheng Ma, XuanQi Zhou}  % 作者，开启标题页才会显示
\affil{\normalsize \textit{School of Artificial Intelligence, Beijing Normal University}}
\date{}  % 日期，开启标题页才会显示
\vspace{-30pt}
\memoto{MCM office}  % 建议书目标
\memofrom{MCM Team 12345678}  % 建议书来源
\memosubject{MCM}  % 建议书主题
\memodate{}  % 建议书日期
\usepackage{setspace}
\titlespacing*{\section}{0pt}{0.5em}{0.5em}  % 调整 section 的间距
\usepackage{titlesec}

\titleformat{\title}
    {\normalfont\huge\bfseries}{\thetitle}{-100em}{}
\titlespacing*{\title}
    {-10pt}{-10pt}{-10pt} % 减小标题和内容之间的间距
    
\begin{document}  % 文档
\thispagestyle{empty}  % 添加这行，使当前页不显示页码
% \vspace{-17pt}
\begin{abstract}  % 摘要
\begin{spacing}{1.1} 

This paper presents a comprehensive framework for time series prediction using a hybrid model that combines \textbf{ARIMA and LSTM}. The model incorporates feature engineering techniques, including \textbf{embedding} and \textbf{PCA}, to transform raw data into a lower-dimensional representation while retaining key information. The embedding technique is used to convert categorical data into continuous vectors, facilitating the capture of complex relationships. PCA is applied to reduce dimensionality and extract principal components, enhancing model performance and computational efficiency. To handle both linear and nonlinear patterns in the data, the ARIMA model captures linear trends, while the LSTM model models complex nonlinear dependencies. The hybrid model is trained on historical data and achieves high accuracy, as demonstrated by low RMSE and MAE scores. Additionally, the paper employs the \textbf{run test} to assess the randomness of sequences, providing insights into the underlying patterns. Ablation studies are conducted to validate the roles of different components in the model, demonstrating the significance of each module. The paper also utilizes the \textbf{SHAP} method to quantify the impact of traditional advantages on the predicted results, offering a detailed understanding of feature importance. The \textbf{KNN} method is used to determine the optimal prediction interval, further enhancing the model's accuracy. The results highlight the effectiveness of combining traditional statistical methods with modern deep learning techniques for robust time series forecasting in Sports.
\end{spacing}

\vspace{10pt}  % 在keywords前添加垂直空间，可以调整数值

  % 结束关键词
\end{abstract}  % 结束摘要
\maketitle  % 生成sheet页

\begin{spacing}{0.5}
\tableofcontents  % 生成目录表
\end{spacing}

%%%%%%%%%%%%%%%%%% sheet页与目录页结束 %%%%%%%%%%%%%%%%%%

\newpage  % 开始新的一页
\section{Introduction}  % 一级标题
\subsection{Background and Rephrasing the question}
The Olympic Games, held every four years and rooted in ancient Greece, promote international friendship and athletic excellence while emphasizing fairness, gender equality, and inclusivity, and our analysis of historical data aims to model national medal tallies to provide insights into future Olympic outcomes.
Based on data, our work is primarily divided into three tasks: 
\begin{enumerate}
    \item Build a prediction model: Construct models to predict each country's medal count (including golds) for the 2028 Los Angeles Summer Olympics, analyzing trends and identifying potential first-time medal winners.
    \item "Great Coach" Effect Analysis: Examine the impact of exceptional coaches on performance, evaluate their influence, and recommend sports for investment in outstanding coaches for three countries chosen.
    \item Decision Support Insights: Analyze additional features revealed by our models to provide recommendations for National Olympic Committees.
\end{enumerate}

\subsection{Overview of Steps}

% 上半部分图
\begin{figure}[htbp]
    \centering
    \includegraphics[width=0.9\linewidth]{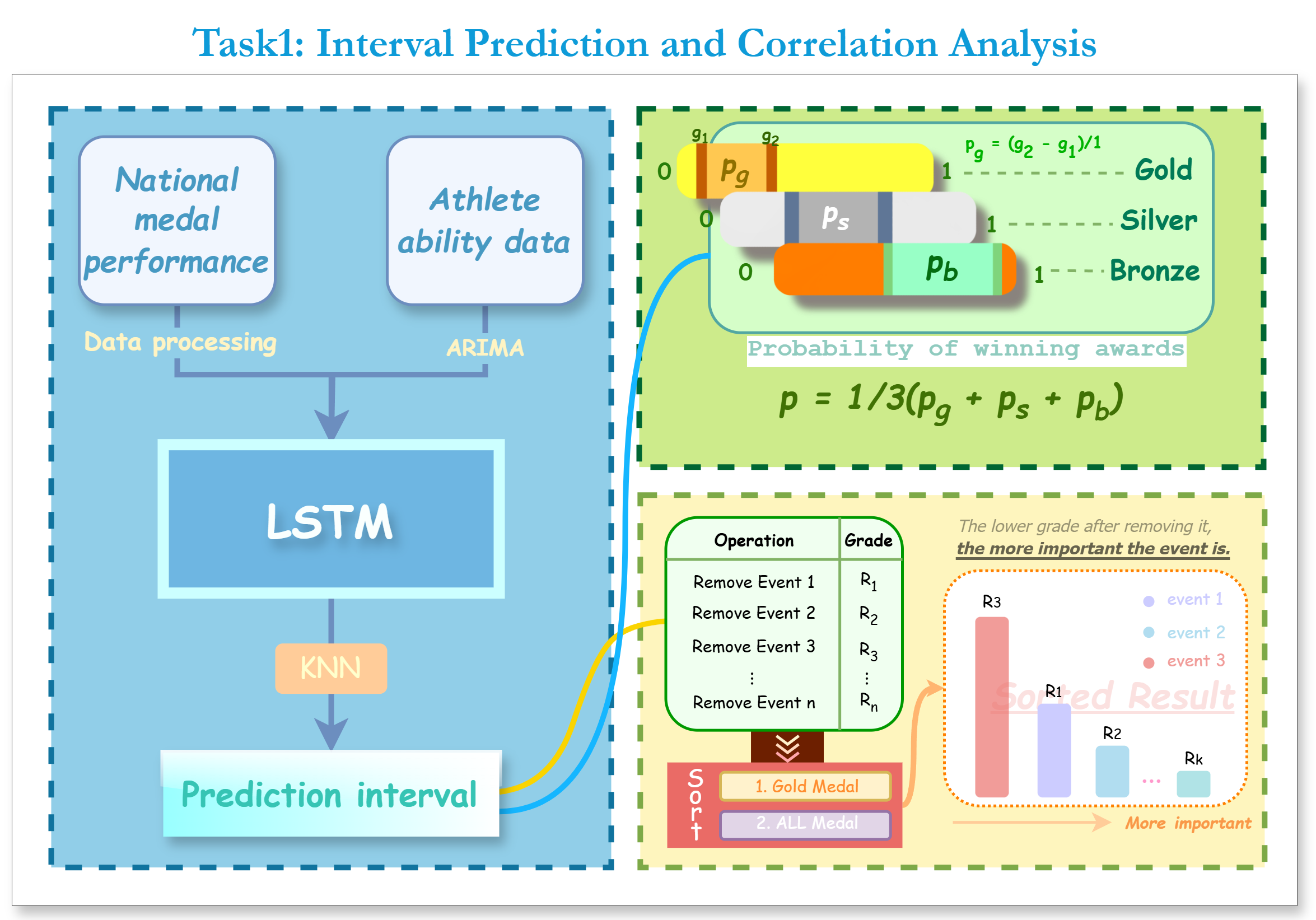}
    \caption{Task 1 Solution Overview}
    \label{fig:step1}
\end{figure}

\newpage  % 或者使用 \clearpage

% 下半部分图
\begin{figure}[htbp]
    \centering
    \includegraphics[width=0.9\linewidth]{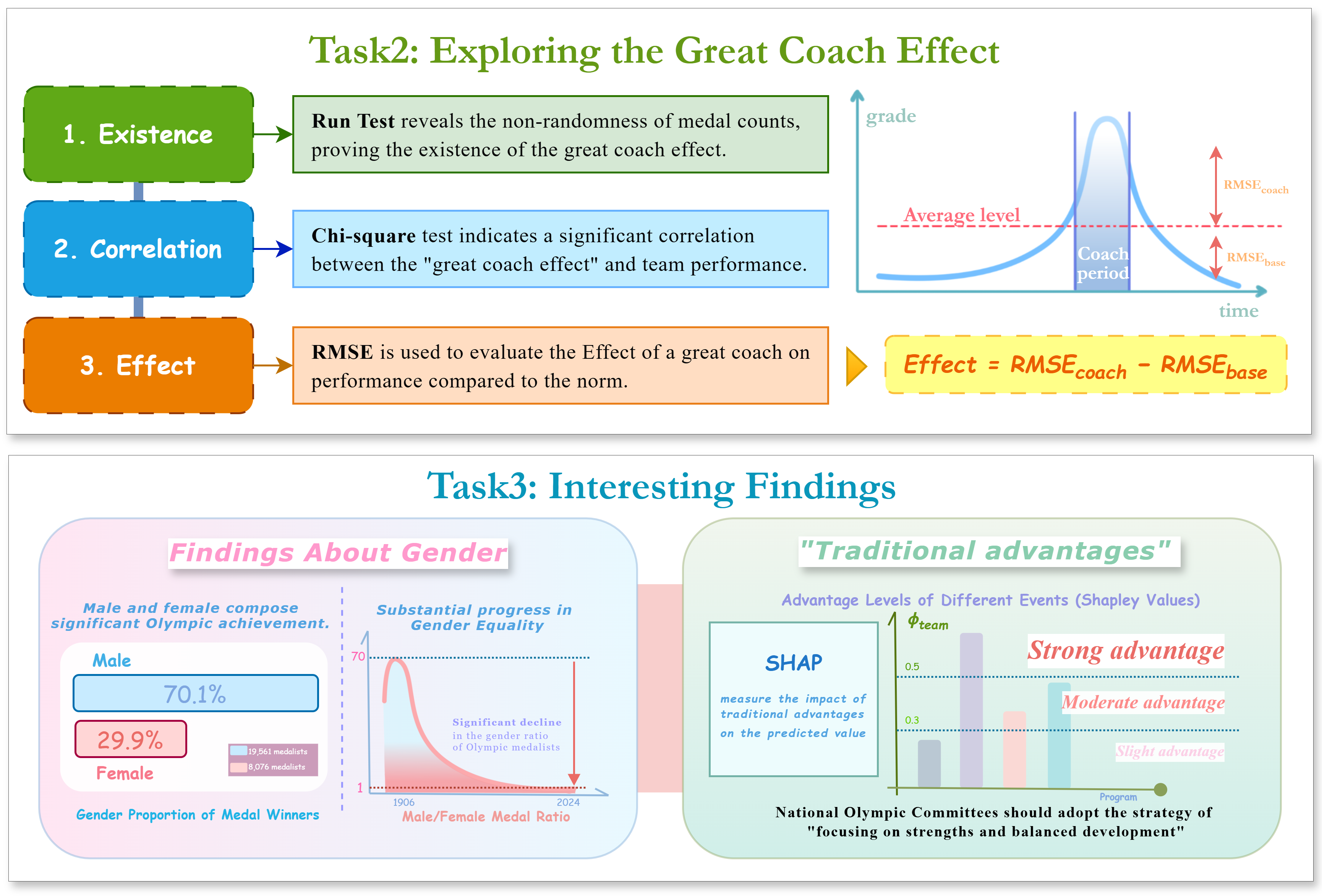}
    \caption{Task 2\&3 Solution Overview}
    \label{fig:step2}
\end{figure}

\label{fig:steps}

% To address these issues, our team will take the following:
% \begin{itemize}  % 无序列表
% \item Perform outlier processing on the data provided by COMAP official.
% \item Analytical coding was conducted for the corresponding data of the athletes for feature selection and pca feature extraction.
% \item For each edition of the Olympics, we organize data by project and arrange them into a matrix $ M_{t-1}$ by sport.
% \item Apply the ARIMA model to conduct time series modeling on data of each country for every edition of the Olympics. 
% \item Using the ARIMA model, we will predict this year's performance $N_{t-1}$
% based on the previous year's results $N_{t-2}$.
% \item At last, we will input processed data above into the LSTM model for training, aiming to learn temporal features from historical data.

% \end{itemize}  % 无序列表结束

\section{ Assumptions and Symbols  }  % 一级标题
\subsection{Model Hypothesis}

To streamline our modeling approach, we make the following assumptions:
\vspace{-0.5em} % Adjust the value to reduce space before the list
\begin{itemize}
    \setlength\itemsep{0em} % Reduce spacing between items
    \item \textbf{Exclusion of Long-Term Athletes:} Athletes participating in more than five consecutive Olympic Games are excluded from the time series state for node $t$ to reduce noise.
    \item \textbf{Focus on Provided Data Factors:} Only factors explicitly mentioned in the problem statement are considered, ignoring unspecified influences for simplicity.
    \item \textbf{Data Reliability with Exceptions:} Data is assumed accurate and reliable, except for a few specific Olympic Games identified as exceptions.
\end{itemize}
\subsection{Symbols and Definitions}
Our definitions of the various types of symbols are located at Table \ref{tab:symbols1}.
\begin{table}[ht]
\centering
\renewcommand{\arraystretch}{1.2} % 调整行间距
\caption{Symbols and Definitions}
\begin{tabular}{@{}cc@{}} % 使用 @{} 去掉左右的多余间距
\toprule
\textbf{Symbol} & \textbf{Definition} \\ 
\midrule
$X_t$ & Observation at time $t$ \\
$\phi_i$ & Autoregressive coefficient for lag $i$ \\
$\epsilon_t$ & Error term at time $t$ \\
$\Delta X_t$ & Differenced time series at time $t$ \\
$\mu$ & Constant term in the moving average model \\
$\theta_i$ & Moving average coefficient for lag $i$ \\
$\mathbf{P}$ & Projection matrix \\
$\mathbf{v}_{\text{new}}$ & Transformed vector \\
$\mathbf{v}_{\text{old}}$ & Original feature vector \\
$\mathbf{v}_{\text{combined}}$ & Combined feature vector \\
$\mathbf{v}_{\text{reduced}}$ & Reduced feature vector \\
$\mathbf{N}_{\text{team}}$ & National team feature matrix \\
$\mathbf{n}_{\text{feature}}$ & Encoded feature vector \\
$d(x_q, x_i)$ & Euclidean distance between $x_q$ and $x_i$ \\
$\rho$ & Spearman rank correlation coefficient \\
$d_i$ & Difference in ranks for Spearman correlation \\
\bottomrule
\end{tabular}
\label{tab:symbols1}
\end{table}

%%%%%%%%%%%%%%%%%%%%%%%% 并排图 %%%%%%%%%%%%%%%%%%%%%%%%

\section{Data Pre-processing}  % 一级标题

Since we could only use the official COMAP dataset ``2025\_Problem\_C\_Data.zip'', which was compiled based on the official IOC website as well as records from other sources, there may be outliers and missing values. Therefore, we pre-processed these data files before modeling.

\begin{itemize}
    \item \textbf{Glitch removal:} We found some unreadable gibberish in the \textit{summerOly\_programs.csv} file and removed it and replaced it with the average of the before and after data.
    
    \item \textbf{Missing information handling:} In the \textit{summerOly\_athletes.csv} file, we removed some entries with large missing information about the athletes.
    
    \item \textbf{Handling of outliers:} In \textit{summerOly\_hosts.csv} and other files, we removed entries for years when the Olympic Games could not be organized due to war.
    
    \item \textbf{Name consistency:} Harmonize country identifiers across files in different data files (e.g., ``United States'' and ``USA'').
\end{itemize}

\section{Task1: Interval Prediction and Correlation Analysis}  % 一级标题
\begin{figure}[htbp]
    \centering    \includegraphics[width=0.8\textwidth]{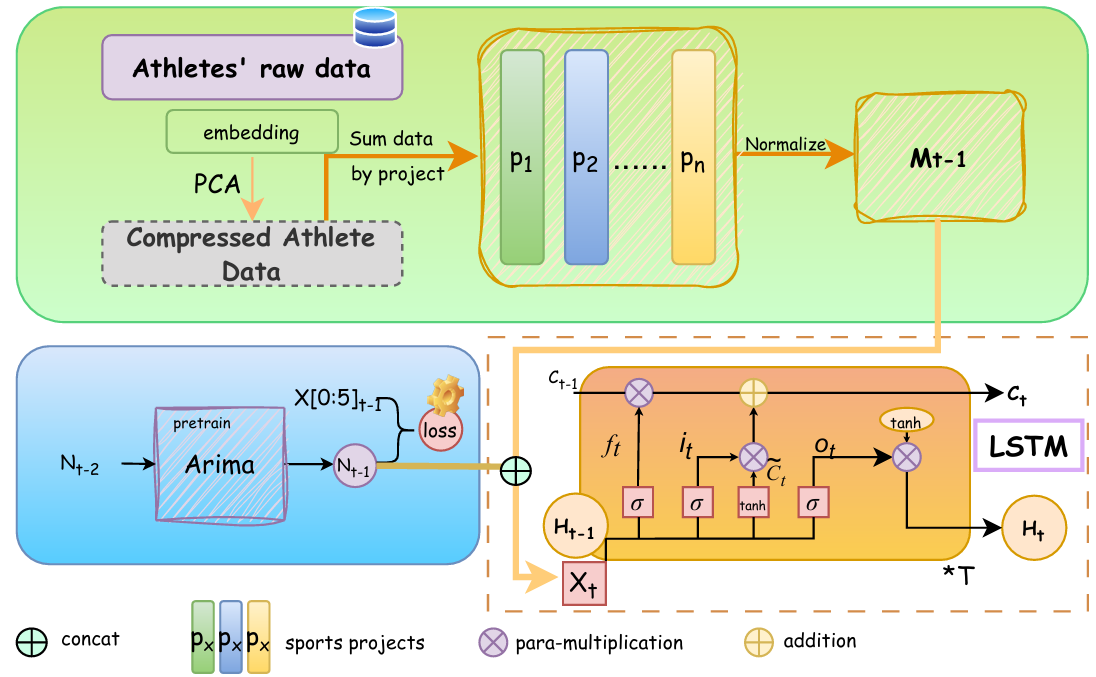} % Adjust the width as needed
    \caption{ARIMA-LSTM Hybrid Network Architecture}
    \label{fig:flow1}
\end{figure}

In order to establish the prediction model of time series, we used the recurrent neural network LSTM, and introduced ARIMA processing to perfect the LSTM imperfect linear sequence capture, constituting a network structure as in Fig. \ref{fig:flow1}., in the processing of the features as network inputs, we carried out the feature extraction, feature dimensionality reduction and feature coding, which greatly solved the problem of the feature scale, in accordance with the time of each country by time to train the iterative We use the MSE calculation as the loss function, back propagation and KNN to get the prediction results, and the experimental data show that our model has good prediction performance and solves the problem well. In addition, we conducted ablation experiments and correlation analysis to verify the stability and robustness of our model in the whole process stage.

\subsection{Characterization of Data}
Analyzing the given data, we can see that the predicted results for the next year are influenced by the change in the number of medals of the country in the past time, the number of athletes involved, the number of sports, the information indicators of athletes of each country in each sport, and so on. For the purpose of modeling, we start by exploring the potential impact of each type of information over time, by treating the information as vintage features or as features within a time step. We summarize the information as shown in Table \ref{tb2} below to better analyze the feature extraction.

\begin{table}[ht]
\centering
\caption{Characterization of Data}
\vspace{2mm}
\label{tb2}
\begin{tabular}{c|c|c} % Adjust the widths as needed
\toprule
\textbf{Athletes} & \textbf{Amount} & \textbf{Element} \\ \hline
NOC & 235 & CHN, DEN, FIN... \\ \hline
edition & 31 & 1, 2, 3... 31 \\ \hline
sport& 71 & Football, Judo, Sailing...\\ \hline
awards & 4 & Gold, Silver, Bronze, No medal \\ \hline
games& 66& Badminton Men's Singles...\\ \midrule[2pt]
\textbf{Non-athletes} & \textbf{Amount} & \textbf{Element} \\ \hline
gold & 83 & [0, 83] \\ \hline
silver & 78 & [0, 78] \\ \hline
bronze & 77 & [0, 77] \\ \hline
athletes & 1109 & [0, 1109] \\ \hline
events & 47 & [0, 47] \\ 
\bottomrule

\end{tabular}
\label{tab:symbols}
\end{table}

\subsection{Feature Embedding and Feature Dimensionality Reduction using PCA}
\subsubsection{Athlete Evaluation Analysis}
The evaluation of athletes is a critical step in making time projections. We aim to uncover distinguishing features of different athletes, which are used  construct parts of the time step. . Feature embedding\cite{wu2024deepfeatureembeddingtabular} is a powerful technique that allows us to transform categorical data into a continuous vector space, making it easier to capture complex relationships and patterns. By embedding features, we can effectively handle categorical variables and reduce dimensionality, which is crucial for improving model performance and computational efficiency.

To identify implicitly varying features, we combined the information given in the question with five categories: NOC, the earliest year of participation in the Olympics, the number of Olympic Games participated in, the best awards received, and the sport in which the athlete participated. Each category, except for gender, was encoded with a seed of 42, resulting in coded vectors with a dimension of 10.

For example, the NOC, which includes 235 types, was encoded into a 1 × 10 vector, denoted as $\mathbf{v}_{\text{NOC}}$. Other categories were similarly encoded into a consistent space, effectively addressing the challenge of unifying different types of data for evaluation. For each athlete, we obtain five 1 × 10 vectors: $\mathbf{v}_{\text{NOC}}$, $\mathbf{v}_{\text{edition}}$, $\mathbf{v}_{\text{games}}$, $\mathbf{v}_{\text{awards}}$, and $\mathbf{v}_{\text{sport}}$. These vectors are combined into a single 1 × 50 vector, denoted as $\mathbf{v}_{\text{combined}}$.

\begin{equation}
\mathbf{v}_{\text{combined}} = \begin{bmatrix} \mathbf{v}_{\text{NOC}} & \mathbf{v}_{\text{edition}} & \mathbf{v}_{\text{games}} & \mathbf{v}_{\text{awards}} & \mathbf{v}_{\text{sport}} \end{bmatrix}_{1\times50}
\end{equation}
To perform dimensionality reduction, we apply Principal Component Analysis (PCA)\cite{shlens2014tutorialprincipalcomponentanalysis} to the combined feature vectors of all athletes. The process begins by calculating the covariance matrix $\mathbf{C}$, which is obtained by summing the outer products of each athlete's feature vector $\mathbf{v}_{\text{combined}}$ with its transpose:

\begin{equation}
\mathbf{C} = \frac{1}{N} \sum_{i=1}^{N} \mathbf{v}_{\text{combined},i} \mathbf{v}_{\text{combined},i}^T
\end{equation}

where $N$ is the total number of athletes. The covariance matrix is then normalized to ensure that the data is centered.

\begin{figure}[htp]
    \centering
    \includegraphics[width=1.0\linewidth]{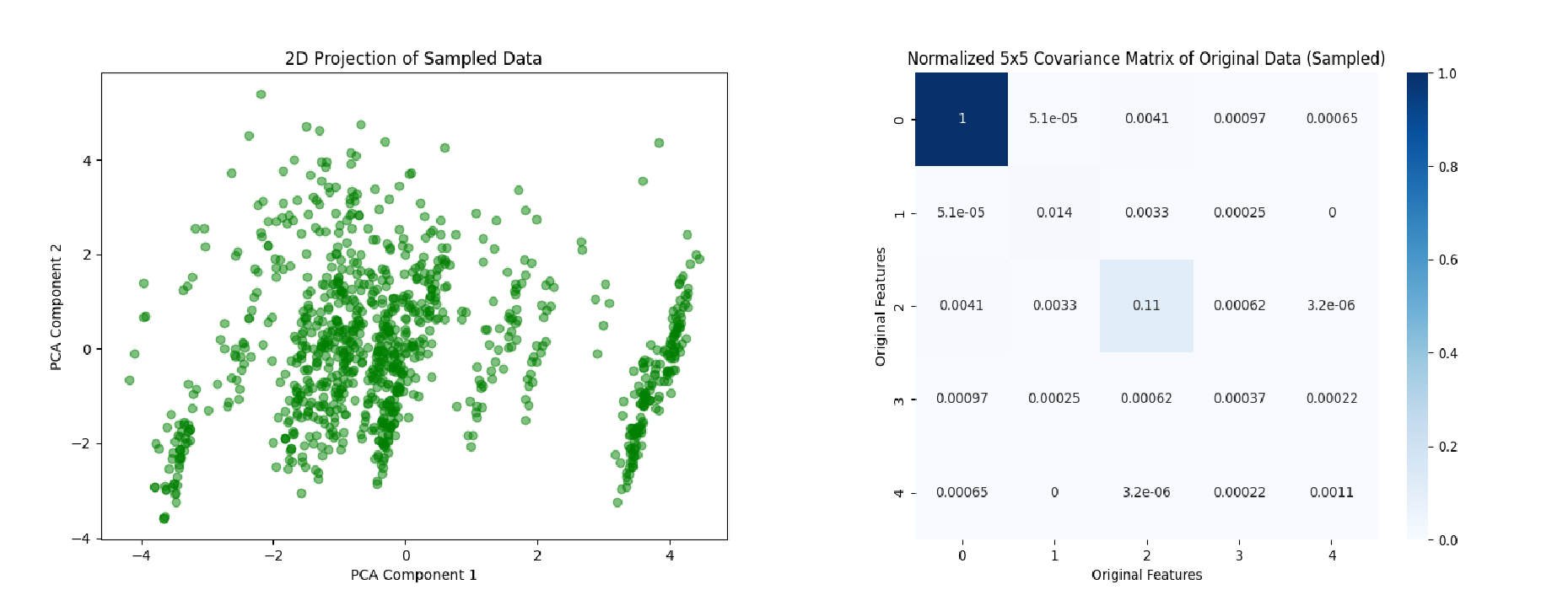}
    \caption{2D projection and covariance matrix of original sampled data}
    \label{fig:pca1}
\end{figure}

The Figure \ref{fig:pca1} shows the two-dimensional projection of the original sampled data (left) and the normalized $5×5$ covariance matrix (right). At this time, the data presents a relatively scattered and overlapping distribution, and the specific feature direction cannot be distinguished.

Next, we perform PCA on the covariance matrix $\mathbf{C}$ to identify the principal components. The process begins with eigenvalue decomposition, where we solve the characteristic equation:

\begin{equation}
\det(\mathbf{C} - \lambda \mathbf{I}) = 0
\end{equation}

Here, $\lambda$ represents the eigenvalues, and $\mathbf{I}$ is the identity matrix. Solving this equation provides the eigenvalues of $\mathbf{C}$. For each eigenvalue $\lambda_i$, we find the corresponding eigenvector $\mathbf{v}_i$ by solving:

\begin{equation}
(\mathbf{C} - \lambda_i \mathbf{I}) \mathbf{v}_i = 0
\end{equation}

These eigenvectors are orthogonal and form the basis of the new feature space. We sort the eigenvalues in descending order and select the eigenvectors corresponding to the largest eigenvalues as the principal components, capturing the most variance in the data.

The selected eigenvectors form the columns of the projection matrix $\mathbf{P}$, which is of size $50 \times 5$:

\begin{equation}
\mathbf{P} = \begin{bmatrix} \mathbf{p}_1 & \mathbf{p}_2 & \mathbf{p}_3 & \mathbf{p}_4 & \mathbf{p}_5 \end{bmatrix}_{50 \times 5}
\end{equation}

Finally, the original feature vectors are transformed using the projection matrix:

\begin{equation}
\mathbf{v}_{\text{reduced}} = \mathbf{P}^T \mathbf{v}_{\text{combined}}
\end{equation}

\begin{figure}[htp]
    \centering
    \includegraphics[width=1.0\linewidth]{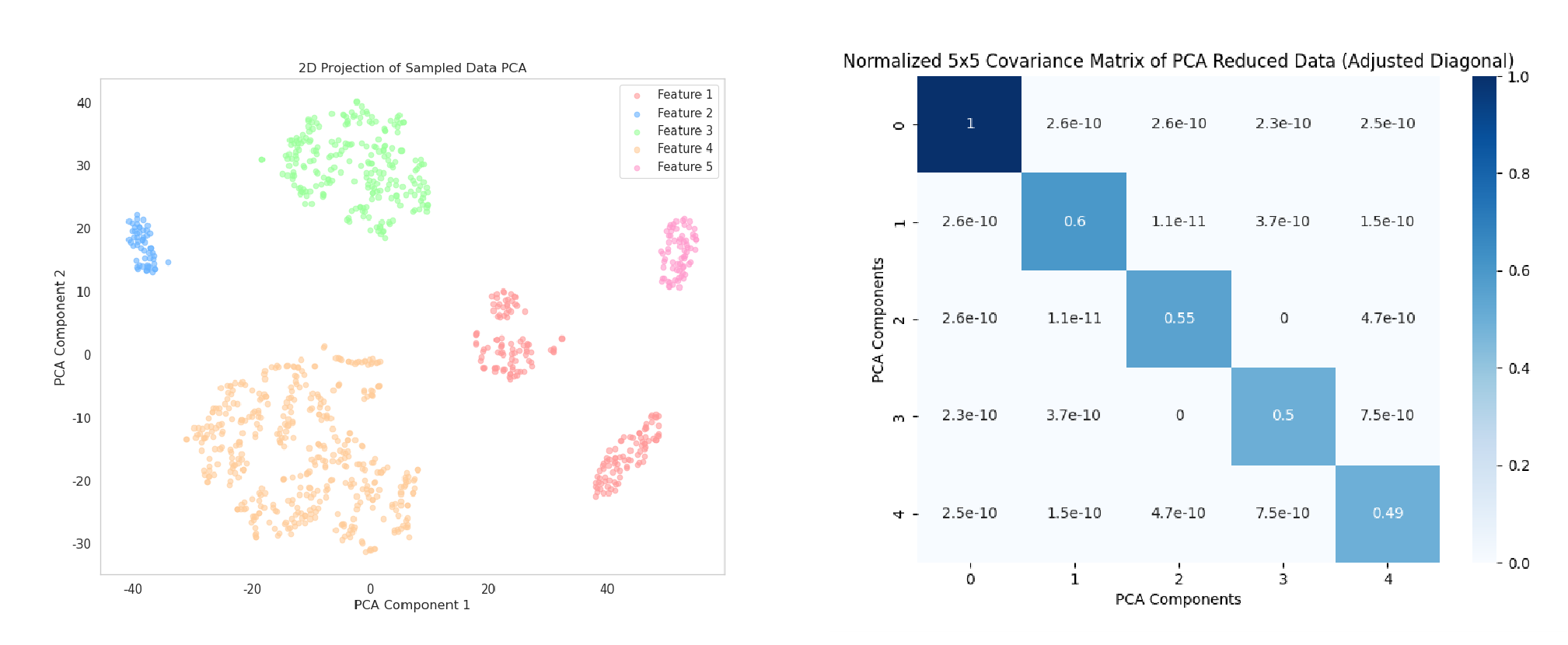}
    \caption{2D projection and covariance matrix of PCA-reduced data}
    \label{fig:pca2}
\end{figure}

The effectiveness of our PCA transformation is shown in Figure \ref{fig:pca2}. The left plot displays the 2D projection of sampled data, highlighting feature clustering. The right shows the normalized covariance matrix of PCA-reduced data, preserving relationships while minimizing noise and redundancy.

This transformation projects the data onto the principal components, retaining the most significant features while reducing noise and redundancy. For the evaluation of athletes, we can currently utilize this to obtain an implicit evaluation vector of an athlete after projective transformation, which plays a role later as a time-step construction.
\subsubsection{National team evaluation analysis (other non-athlete related factors)}
For those non-athlete metrics, which possess significant vintage and separateness between countries and are an important component in constructing the time step, we take a different approach from athlete profiling. In this module, we simply code the features into a uniform spatial dimension. These features include the number of gold, silver, and bronze medals achieved by the country at time $t$, the number of athletes participating, and the number of events they were able to participate in. Each of these quantitative values is encoded as a $1 \times 10$ vector to unify the dimensions, following the methodology used for constructing the embedding in the previous section.

For a given national team at a given point in time, these encoded features form a matrix $\mathbf{N}_{\text{team}}$ of size $10 \times 5$:

\begin{equation}
\mathbf{n}_{\text{team}} = \begin{bmatrix}
\mathbf{n}_{\text{gold}} & \mathbf{n}_{\text{silver}} & \mathbf{n}_{\text{bronze}} & \mathbf{n}_{\text{athletes}} & \mathbf{n}_{\text{sports}}
\end{bmatrix}_{10 \times 5}
\end{equation}

where each row vector $\mathbf{n}_{\text{feature}}$ represents a $1 \times 10$ encoded vector for the respective feature: gold medals, silver medals, bronze medals, number of athletes, and number of events.

\subsection{Introduction to Time-Step Data using sliding window algorithm}\label{subset:timestep_intro}
After coding, we can see that for the evaluation of the athletes, it seems that we have only performed a dimensionality reduction in the evaluation features. To apply them consistently to the time period of the LSTM, we need to manipulate them further. One of the simplest ways to think about this is to project the athletes directly to the corresponding program at the corresponding time $t$, i.e., to accumulate to one row of a matrix representing a total of 73 programs until all athletes of the country have participated in the evaluation.

However, it is clear that this has the problem that some legendary athletes from earlier years will have a high impact but have a 0 probability of competing in the following year. Thus, we use a sliding window to optimize the participation of athletes from this country in the evaluation matrix at a given time, avoiding a large amount of noise. The complete procedure is as follows:

1. Read Athlete Data Vectors: For each athlete, read their data vector of size $1 \times 50$ at a given time before the current Olympics.

2. Project to New Vector: Use the projection matrix $\mathbf{P}$ to transform the $1 \times 50$ vector into a new $1 \times 5$ vector:

   \begin{equation}
   \mathbf{v}_{\text{new}} = \mathbf{v}_{\text{old}} \cdot \mathbf{P}
   \end{equation}

3. Find Corresponding Program: Look up the dictionary to find the corresponding program number for the athlete.

4. Accumulate to Matrix Row: Add the new vector $\mathbf{v}_{\text{new}}$ to the corresponding row of the matrix that represents the program. If the athlete participates in multiple programs, ensure that the data vector is added to each corresponding row.

5. Exclude Outdated Athletes: Exclude athletes who have not participated in the last five Olympic Games or more. These athletes should not influence the evaluation at the current time. Use a sliding window to ensure that only relevant athletes are included.

By following these steps, we ensure that the evaluation matrix accurately reflects the current and relevant contributions of athletes, minimizing noise and enhancing the predictive power of the LSTM model. The processed vectors form part of the state matrix for the national team at a given time, with dimensions $71 \times 5$($M_t$). This matrix is combined with the ARIMA method result matrix from the previous time step for the national team, as described in Section~\ref{subsec:arima}, which has dimensions $10 \times 5$($N_t$), and the host country annotation information.

The host information is a binary vector added as the last row, indicating whether the next time step is a host country (with states $(0, 0, 0, 0, 0)$ or $(1, 1, 1, 1, 1)$). Together, these components form the input state matrix $\mathbf{X}_t$ (\( X_t = M_t \oplus N_t \)) for the LSTM in Section~\ref{subsec:LSTM}.

% Ensure you have labels in your document for these sections:
% \section{ARIMA Method} \label{sec:arima}
% \section{LSTM Input State Matrix} \label{sec:lstm}

\subsection{ARIMA-LSTM Model}
Given the time series nature of the data, after thorough consideration, we have ultimately decided to employ the ARIMA model to capture linear relationships within the data and select the LSTM model to capture non-linear relationships. 

By combining the strengths of both models—the ARIMA model's proficiency in handling linear time series data for effective modeling and forecasting of sequences with stable trends and seasonality\cite{5953012}, and the LSTM model's capability as a type of recurrent neural network (RNN) to capture complex patterns and non-linear dependencies over long periods—this approach aims to provide a more robust and accurate prediction framework. 

\subsubsection{ARIMA model}\label{subsec:arima}
ARIMA (Autoregressive Integrated Moving Average) is a widely used method for time series analysis. The ARIMA($p, d, q$) model consists of three key parameters\cite{ijert2021modeling}: the autoregressive order $p$, the differencing order $d$, and the moving average order $q$, where d represents the number of differences required for stationarity, while $p$ and $q$ denote the orders of autoregressive and moving average terms, respectively.

\textbf{Step 1. Parameter Determination  (determine the parameter $p, d, q$)}

\begin{figure}[ht]
    \centering
    \includegraphics[width=1\textwidth]{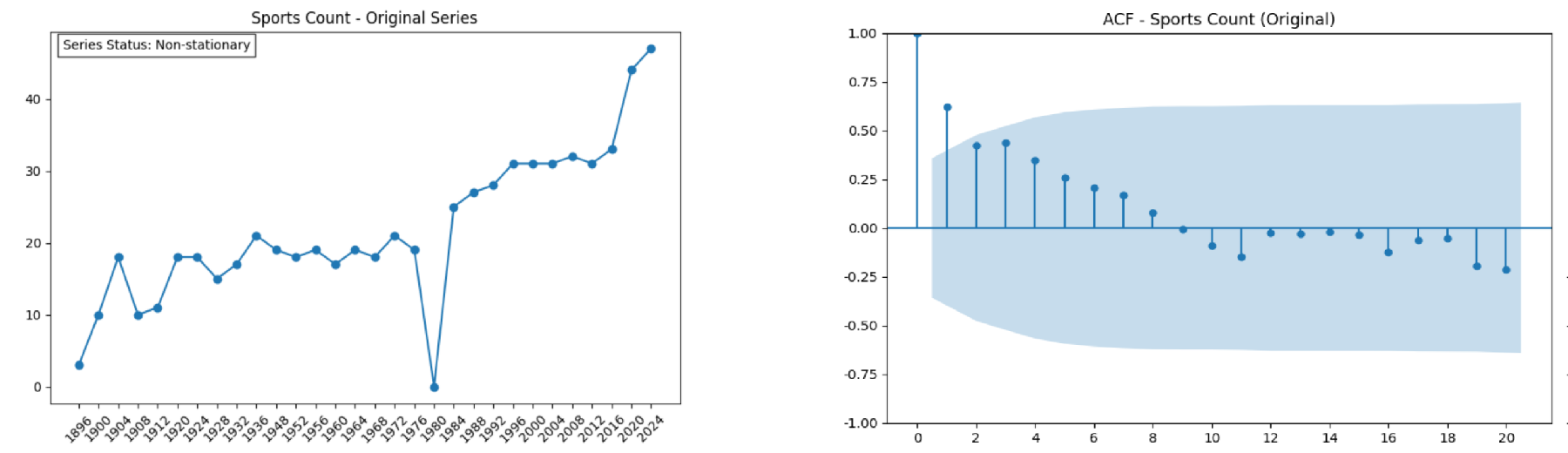} % Adjust the width as needed
    \caption{Number of entries and ACF chart over the years}
    \label{fig:arima11}
\end{figure}

First, we conducted smoothness tests on each feature $f$ Taking the number of projects participated by each country as an example, Figure~\ref{fig:arima11} shows that this time series exhibits a clear trend. Furthermore, we performed unit root tests and found that the series contains a unit root. Therefore, we can conclude that the project number feature $f$  is a non-stationary sequence that requires further smoothing processing.

\begin{figure}[ht]
    \centering
    \includegraphics[width=1\textwidth]{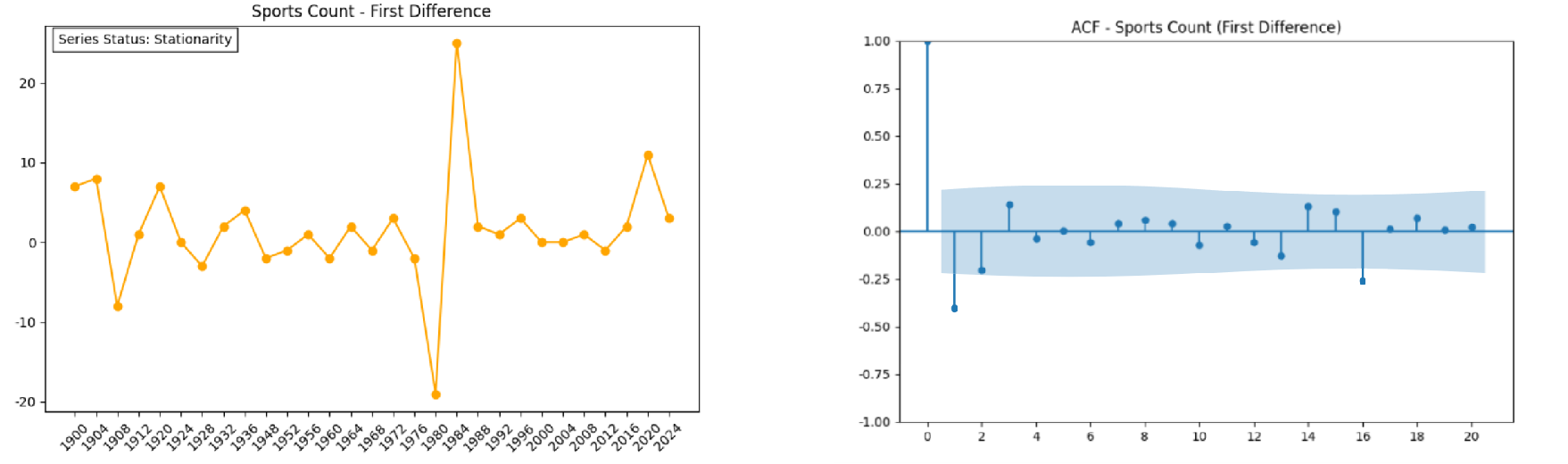} % Adjust the width as needed
    \caption{First-order difference time series and ACF plot}
    \label{fig:arima12}
\end{figure}

Let $\Delta_f = f_t - f_{t-1}$. We further perform a smoothness test on $\Delta_f$. As shown in Figure ~\ref{fig:arima12}, the differenced sequence fluctuates randomly around a certain value without a significant trend. In addition, the p-values of the unit root test converge to 0, indicating the absence of a unit root. Thus, $\Delta_f$ is confirmed to be a stationary sequence. Other features exhibit similar behavior to $f$, therefore the overall differencing order $d = 1$ is adopted.

For a differenced time series, the ARIMA model is expressed as:

\begin{equation}
    \phi(B)(1-B)^d X_t = \theta(B)\varepsilon_t
\end{equation}

Where $\phi(B) = 1 - \phi_1 B - \cdots - \phi_p B^p $ and $\theta(B) = 1 + \theta_1 B + \cdots + \theta_q B^q$ are the autoregressive and moving average polynomials respectively, and B is the backshift operator. By analyzing the Autocorrelation Function (ACF) and Partial Autocorrelation Function (PACF) plots, the preliminary ranges for p and q can be determined: the truncation in the PACF indicates the order of the autoregressive parameter p, while the truncation in the ACF indicates the order of the moving average parameter q. Subsequently, the Akaike Information Criterion (AIC) or Bayesian Information Criterion (BIC) is employed to evaluate the model's performance across different parameter combinations. The optimal values of p and q are the parameter combinations that minimize these information criteria, thereby achieving a balance between model fit and complexity.

\textbf{Step 2. Pre-trained ARIMA model }

The mathematical expression for the ARIMA($p,d,q$) model is as follows:
\begin{equation}
    X'_t = c + \sum_{i=1}^p \phi_i X'_{t-i} + \sum_{j=1}^q \theta_j \epsilon_{t-j} + \epsilon_t
\end{equation}
 where $X'_t$ represents the differenced time series, $c$ is the constant term, $\phi_i$ denotes the autoregressive coefficients, $\theta_j$ represents the moving average coefficients, and $\epsilon_t$ is the white noise term. Clearly, the parameters to be trained include the following:

1. Autoregressive coefficients \(\phi_1, \phi_2, \ldots, \phi_p\): These capture the strength of the relationship between the time series and its lagged values.

2. Moving average coefficients \(\theta_1, \theta_2, \ldots, \theta_q\): These represent the relationship between the time series and past prediction errors.

3. Constant term \(c\): This reflects the overall level or drift of the time series.

To achieve accurate predictions of national teams' performance characteristics in successive Olympic Games, the feature vectors of each country in each Olympic event, denoted as \(\mathbf{N}_{\text{team}}\), are combined into a feature matrix \(\mathbf{M}_{\text{team}}\), which serves as the model input. Specifically, after filtering, the number of valid Olympic Games is 30, resulting in the feature matrix \(\mathbf{M}_{\text{team}}\) having dimensions \(30 \times 50\) for each country. During the prediction process, the feature vector at the current time step \(t\), \(\mathbf{N}_t\), is predicted using the feature matrix \(\mathbf{M}\), constructed from past time steps (\(0 \sim t-1\)), as the temporal input. The loss function is designed based on the mean squared error (MSE) (Equation:\ref{eq:mse}) between the true feature vector \(\mathbf{N}_{t}^{\text{truth}}\) and the predicted feature vector \(\mathbf{N}_t\).

\textbf{Step 3. Timestep prediction }

After completing the parameter training of the ARIMA model, for the input feature vector \( N_{t-2} \) , the time step prediction of \( N_{t-1} \)  can be performed. For single-step prediction at time point  \( t-1 \), the prediction equation can be expressed as:
\begin{equation}
    \hat{X}_{t-1} = c + \sum_{i=1}^{p} \phi_i X_{t-1-i} + \sum_{j=1}^{q} \theta_j \epsilon_{t-1-j}
\end{equation}

where \( \hat{X}_{t-1} \) is the predicted value at time \( t-1 \), \( X_{t-1-i} \) represents the historical observed values, \( \epsilon_{t-1-j} \) denotes the historical residual terms, and \( c \) is a constant term.
Subsequently, we evaluate the prediction accuracy using the actual features\( X_{t-1} \)  and employ the Root Mean Square Error (RMSE)  (Equation:\ref{eq:mse}) as the metric to assess the prediction performance.

Finally, we concatenate the predicted features \( N_{t-1} \) with the athlete feature matrix to serve as the input for the LSTM network.

\subsubsection{LSTM model}\label{subsec:LSTM}
 The Long Short-Term Memory (LSTM) network\cite{hochreiter1997long} is a specialized type of Recurrent Neural Network (RNN) designed to address the challenges traditional RNNs face in handling long-term dependencies. By introducing a "cell state" and three gating mechanisms—the forget gate, input gate, and output gate—LSTM effectively manages the long-term storage and flow of information\cite{unknown2}. This design allows LSTMs to learn long-term dependencies in time series data, thereby better capturing complex patterns within the data.

The core component of an LSTM is the memory cell, which is responsible for storing and transmitting information. Each memory cell is equipped with three gating mechanisms to control the flow of information----Forget Gate, Input Gate, and Output Gate.

Assuming $ x_t $ is the current input vector, $ h_{t-1} $ is the hidden state from the previous time step, and $ c_{t-1} $ is the cell state from the previous time step, the computation process of an LSTM unfolds as follows:

1. Forget Gate:
The forget gate uses a sigmoid layer to decide what information we should discard from the cell state. It takes as input the previous hidden state $ h_{t-1} $ and the current input $ x_t $ , producing a vector $ f_t $ of values between 0 and 1, where 1 means completely retain, and 0 means completely discard.
\begin{equation}
f_t = \sigma(W_f \cdot [h_{t-1}, X_t] + b_f)
\end{equation}

2. Input Gate:
The input gate consists of two parts:
A sigmoid layer that decides what information to update;
A tanh layer that creates a new candidate vector $ \tilde{c}_t $ to potentially add to the cell state.  
\begin{equation}
i_t = \sigma(W_i \cdot [h_{t-1}, X_t] + b_i)
\end{equation}
\begin{equation}
\tilde{c}_t = \tanh(W_c \cdot [h_{t-1}, X_t] + b_c)
\end{equation}

3. Cell State Update:
The new cell state $ c_t $ is updated from the old state $ c_{t-1} $ by deciding what to forget via the forget gate and selectively adding new information through the input gate.
\begin{equation}
c_t = f_t \odot c_{t-1} + i_t \odot \tilde{c}_t
\end{equation}

Where $ \odot $ denotes element-wise multiplication.

4. Output Gate:
Finally, the output gate determines the final output. It first passes through a sigmoid layer to decide which part of the cell state should be output and then multiplies this result with the cell state passed through a tanh function, yielding the final hidden state $ h_t $. 
\begin{equation}
o_t = \sigma(W_o \cdot [h_{t-1}, x_t] + b_o)
\end{equation}
\begin{equation}
h_t = o_t \odot \tanh(c_t)
\end{equation}

\subsubsection{Loss Function}
\textbf{1. The Mean Absolute Error (MAE):} The Mean Absolute Error (MAE) measures the average difference between predicted values and actual values, with smaller values indicating more accurate predictions. Its unit is the same as that of the target value, allowing direct comparison with the actual magnitude of the data.\cite{unknown} Compared to MSE and RMSE, MAE is less sensitive to outliers and does not amplify results due to the square of a few large errors, thus better reflecting the overall trend of the data.
For actual scores $ A $ and predicted scores $ \hat{A} $, MAE can be calculated using the following formula:

\begin{equation}
\text{MAE} = \frac{1}{mn} \sum_{i=1}^{m} \sum_{j=1}^{n} |A_{ij} - \hat{A}_{ij}| 
\end{equation}

Here, $ A_{ij} $ and $ \hat{A}_{ij} $ represent the elements at position $ (i,j) $ in matrices $ A $ and $ \hat{A} $, respectively, where $ m $ and $ n $ denote the number of rows and columns of the matrices. MAE is the average of the absolute differences between corresponding elements.

\textbf{2. Mean Squared Error (MSE):} MSE is commonly used to assess the difference between predicted and actual values. It quantifies the prediction performance by calculating the mean of squared errors between predictions and actual values. By squaring the error terms, MSE assigns greater weight to larger errors. Thus, MSE is highly sensitive to significant errors, making it effective for highlighting substantial prediction inaccuracies.
The calculation method for MSE is as follows:

\begin{equation}
\text{MSE} = \frac{1}{mn} \sum_{i=1}^{m} \sum_{j=1}^{n} (A_{ij} - \hat{A}_{ij})^2 
\label{eq:mse}
\end{equation}

\textbf{3. Root Mean Squared Error (RMSE):} RMSE is the square root of MSE, retaining the property of amplifying larger errors while converting the result back to the original data units, making it easier to interpret. Therefore, RMSE is frequently used when reporting the final model performance.

\begin{equation}
\text{RMSE} = \sqrt{\frac{1}{mn} \sum_{i=1}^{m} \sum_{j=1}^{n} (A_{ij} - \hat{A}_{ij})^2} 
\label{eq:rmse}
\end{equation}
% \begin{figure}[htp]
%     \centering
%     \includegraphics[width=0.8\linewidth]{figures/loss.png}  % 将宽度调整为0.5
%     \caption{Loss when LSTM(RMSE)}
%     \label{fig:loss}
% \end{figure}
\subsection{Finding the Optimal Interval Using KNN}

\begin{figure}[htp]
    \centering
    \includegraphics[width=0.4\linewidth]{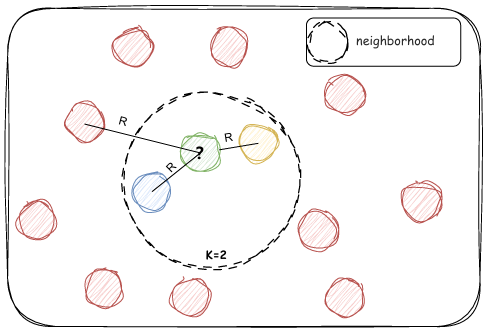}  % 将宽度调整为0.5
    \caption{knn method}
    \label{fig:knn}
\end{figure}

The K-Nearest Neighbors (KNN) algorithm is a simple yet effective supervised learning method widely used for classification and regression tasks. The core idea is to compute the distance between the sample to be predicted and all samples in the training set, select the K nearest neighbors, and determine the class or value of the sample to be predicted based on the labels or values of these neighbors.\cite{9065747}We draw Figure \ref{fig:knn} to show KNN vividly.

In our work, the raw data is transformed into a vector space through embedding encoding. We will utilize the K-Nearest Neighbors (KNN) algorithm to calculate the Euclidean distances between the prediction target and the sample points in this vector space, and select the two nearest neighbors as the final prediction interval.

The Euclidean distance calculation formula is as follows:
\begin{equation}
d(x_q, x_i) = \sqrt{\sum_{k=1}^{n} (x_{qk} - x_{ik})^2}
\end{equation}
where $ x_q $ is the prediction target, and $ x_i $ is a sample point.

We will select the performances of the two sample points with the smallest Euclidean distances to the prediction target as the predicted interval. By using this method, we can estimate the likely range of the prediction target based on the performance of its nearest neighbors.
\subsection{Our Task1 Results }
\begin{figure}[htp]
    \centering
    \includegraphics[width=0.8\linewidth]{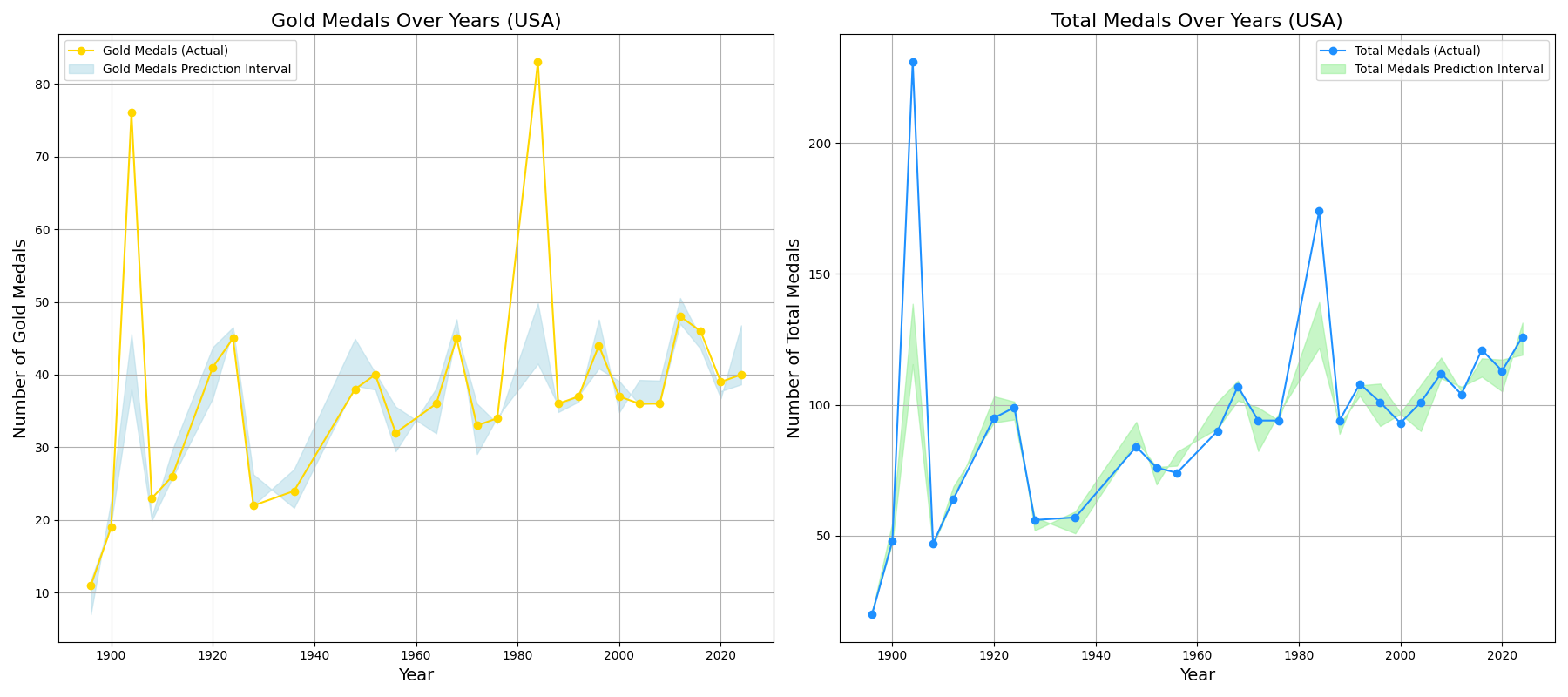}  % 将宽度调整为0.5
    \caption{Prediction Results}
    \label{fig:ayear}
\end{figure}

\begin{figure}[htp]
    \centering
    \includegraphics[width=0.7\linewidth]{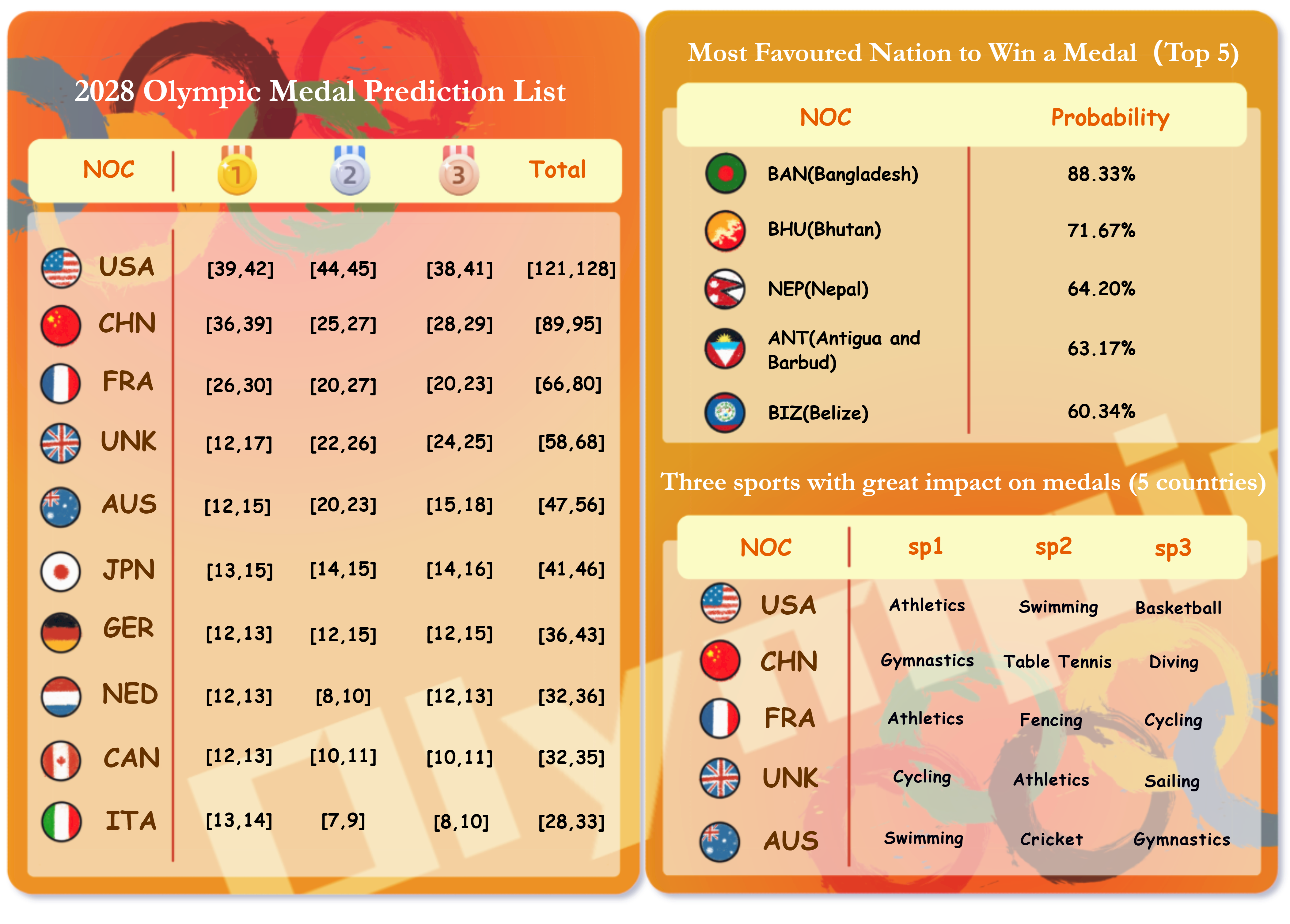}
    \caption{Task1 Results}
    \label{fig:results_1}
\end{figure}

For our arima-lstm training(epoches=500), the best model we trained got \textbf{RMSE}:0.098 and \textbf{MAE}:0.072, which gave very good results in the ten-dimensional embedding space.
For the first problem, after our model generates the matrix and then maps it to the quantity interval by KNN, we can predict the number of medals (gold, silver, and bronze) of the country at each time. In order to demonstrate this visually, we select the top ten countries in terms of the total number of medals and show the prediction results of our model as shown in Figure \ref{fig:results_1}. At the same time, to reflect the effect of our model on temporal extrapolation, we will show the predicted change in gold medals versus total medals for the United States over time variations and compare it to the true value, as shown in Figure \ref{fig:ayear}.Based on analysing the data, we find that the top three countries most likely to increase their medals are CAN (6.5), GER (5) and FRA (4), and the countries most likely to get worse in terms of medal counts are  KOR(-10), ITA (-9.5), HUN (-6).
For the second problem, we specifically analyzed countries that have never won a medal, mapping the results of the KNN (the ratio of the distances of the final resulting vectors to the vectors of 0 and 1 in the embedding space) to probabilities via Logistic Regression. We consider countries with this probability greater than 0.5 as promising to win the first medal. We ended up finding twelve countries for which their NOCs are BAN, BHU, NEP, ANT, BIZ, CAM, GBS, MAW, MTN, NCA, TLS, TGA. We present the five countries with the highest probability as shown in Figure \ref{fig:results_1}.

For the third problem, we analyzed the LSTM predicted 2028 national team athletes' total state matrix, compared the size of the corresponding eigenparadigm for each item, and sorted them. We conducted item analysis for different countries and listed the three most important items for medal acquisition for five countries as shown in Figure \ref{fig:results_1}. For all the items across all countries, we statistically found that athletics, swimming, gymnastics, shooting, and cycling are the top five items that most affect the number of medals, and also the top five items with the highest feature importance. 
For the host country's selected events, we analyze the project impact brought by the host by adjusting the LSTM's prediction of the previous time step. We find that in the host's selected events, the predicted average number of medals gained per category of the selected events increases by 4.36\%..Among these, the five events that have the highest host effect on the project, leading to the highest total medals gain, are Athletics (predicted increase of 12.33\%), Swimming (12.13\%), Cycling (9.52\%), Fencing (8.97\%), and Basketball (8.81\%).


\subsection{Correlation Analysis and Ablation Study }
\subsubsection{Spearman Correlation Coefficient}

The Spearman correlation coefficient is obtained by calculating the ranks of two variables $ x $ and $ y $. Specifically, the steps are as follows: First, sort the variable $ x = \{x_1, x_2, \ldots, x_n\} $ in ascending or descending order to obtain the sorted sequence $ a = \{a_1, a_2, \ldots, a_n\} $, where the position of each element $ x_i $ in the sequence $ a $ is denoted as $ r_i $, which is the rank of $ x_i $. Similarly, sort the variable $ y = \{y_1, y_2, \ldots, y_n\} $ to obtain its rank sequence $ s = \{s_1, s_2, \ldots, s_n\} $. Then, compute the difference sequence $ d = \{d_1, d_2, \ldots, d_n\} $ between the rank sequences $ r $ and $ s $ (where $ d_i = r_i - s_i $), and substitute it into the Spearman rank correlation coefficient formula:

\begin{equation}
\rho = 1 - \frac{6 \sum_{i=1}^{n} d_i^2}{n(n^2 - 1)}
\end{equation}

Here, $ n $ is the sample size, and $ \rho $ is the Spearman rank correlation coefficient between the variables.
In our predictions, we conducted a Spearman correlation coefficient analysis on the median of the gold medal prediction intervals over time and the median of the total medal count changes. We found that there is a significant relationship between the changes in gold medal predictions and the changes in total medal counts, with a calculated Spearman correlation coefficient of \textbf{0.76}. This result reflects a strong positive correlation, indicating that as the predicted number of gold medals increases, the total number of medals also tends to increase. This suggests that improvements in the performance of a country's athletes in gold medal events are likely to be associated with overall enhancements in their medal-winning capabilities across all events.
\subsubsection{Results of ablation study}
To validate the roles of various modules in the model, we designed an ablation experiment. We removed the ARIMA model inference module from the original model and trained the LSTM using the original data in the national team evaluation matrix (non-athlete matrix). We obtained the final converged results of RMSE and MAE values with consistent training cycles, as shown in Table \begin{table}[htpb]
\centering
\renewcommand{\arraystretch}{1.2}
\caption{Model Performance Comparison}
\label{tab:100}
\begin{tabular}{ccc}
\toprule
  & \textbf{RMSE}  & \textbf{MAE}  \\ \hline
\textbf{Original Model Results} & 0.098 & 0.072 \\  % 请替换 0.XX 为实际值
\textbf{Results without ARIMA Module} & 0.121 & 0.094 \\  % 请替换 0.XX 为实际值
\bottomrule
\end{tabular}
\end{table}
From the results, it can be seen that the ARIMA model plays a significant role in capturing the linearity of the data and ensuring its stationarity, which ultimately affects the final predictions. 
\section{Task2: Exploring the Great Coach Effect}  % 一级标题

\subsection{Examining the existence of the Great Coach Effect}

To verify the great coach effect, we examined historical data of national teams with coaching changes, focusing on the U.S. women's gymnastics team's Olympic medal counts. Using the Runs Test, we analyzed whether their medal sequence exhibited non-random patterns, thereby evaluating the coach's influence on team performance.

\subsubsection{Assess the randomness of medal counts using the runs test}

The runs test is a non-parametric statistical method used to determine whether a binary sample (e.g., $X$ and $Y$) originates from a binomial distribution\cite{Baringhaus2016RevisitingTT}. In this test, a sample is drawn from a population containing $X$ and $Y$, with $n$ instances of $X$ and $m$ instances of $Y$. These elements are arranged in the order of sampling to form a sequence. Consecutive identical elements in the sequence are referred to as "runs," and the total number of runs is denoted by $r$.

For this problem, we propose two hypotheses:

\textbf{Null Hypothesis} ($H_0$): The arrangement of above-average (denoted as 1) and below-average (denoted as 0) medal counts for this national team is random, with probability $p$ of occurrence for 1 and $1-p$ for 0.

\textbf{Alternative Hypothesis} ($H_1$): The arrangement of above-average and below-average medal counts is non-random, indicating a pattern in the probability distribution of high and low performance levels.

To conduct the randomness test, we converted the team's medal performance into binary values (0,1). We employed mean-value encoding: assigning 1 to years with medal counts above the arithmetic mean $\bar{x}$, and 0 otherwise.

\begin{equation}
V(x) = \begin{cases}
1, & \text{if } x > \bar{x} \\
0, & \text{otherwise}
\end{cases}
\end{equation}

Then, we can get the team performance sequence as shown in Figure \ref{fig:run_test}:

\begin{figure}[htbp]
    \centering
    \includegraphics[width=0.8\linewidth]{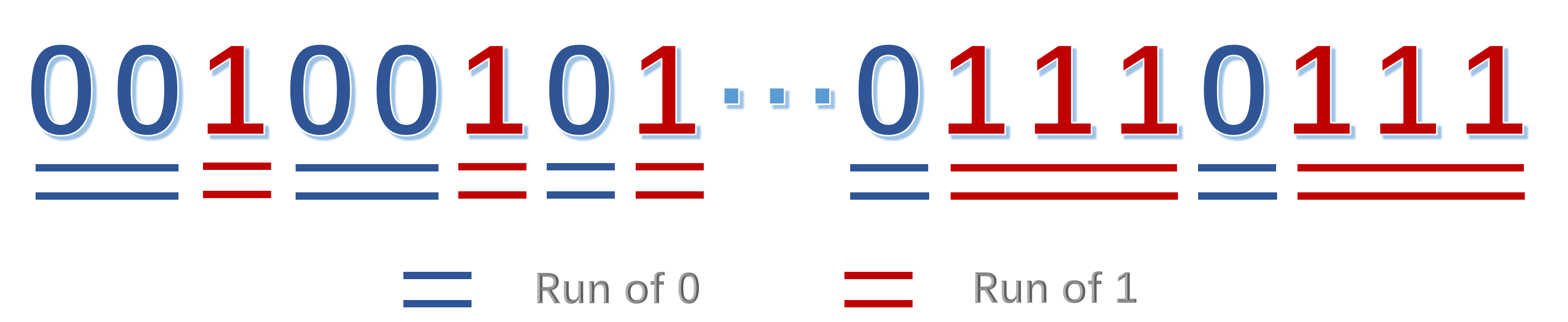}
    \caption{Gymnastics Team Performance Sequence}
    \label{fig:run_test}
\end{figure}

After obtaining the binary sequence, we calculated the actual number of runs $r$, expected number of runs $E(r)$, and variance $V(r)$. The expected number of runs is calculated as:

\begin{equation}
    E(r)=\frac{2n_1n_2}{n_1+n_2}+1
\end{equation}

where $n_1$ and $n_2$ represent the counts of 1s and 0s in the sequence, respectively. By calculating the Z-statistic and corresponding P-value, we can determine the sequence's randomness at a given significance level.

Results show that $E(r)=10$ and $V(r)=4.2353$. The calculated Z-statistic is -1.9437, with a corresponding P-value of 0.0519. At a 90\% confidence level ($\alpha=0.10$), since the P-value is less than the significance level, we reject the null hypothesis, indicating that the sequence exhibits non-random characteristics.

\subsubsection{Coach-Performance Contingency Analysis}

In the preceding section, we employed runs test to examine whether the total medal counts of the national gymnastics team exhibited randomness. The results indicated that the medal counts across different Olympic Games were non-random. Subsequently, to investigate whether the Great Coach Effect correlates with medal acquisition, we conducted a contingency table analysis between medal counts and coaching circumstances.

\begin{table}[htpb]
\centering
\renewcommand{\arraystretch}{1.2}
\caption{Coaching period vs Performance}
\label{tab:5.2.1}
\begin{tabular}{ccc}
\toprule
  & \textbf{Good Performance}  & \textbf{Bad Performance}  \\ \hline
\textbf{Pre-great coach} & $n_{11}$ & $n_{12}$ \\
\textbf{Post-great coach} & $n_{21}$ & $n_{22}$ \\ 
\bottomrule
\end{tabular}
\end{table}

In the table \ref{tab:5.2.1}, $n_{ij}$ represents the frequency with which events in row $i$ and column $j$ occur simultaneously, where $i$ represents the presence or absence of an eminent coach, and $j$ represents the team's performance level. Let the null hypothesis $ H_0 $ indicate that there is no correlation between the coach effect and performance, and the alternative hypothesis $H_1$ indicate that there is a correlation between the coach effect and performance.

Assuming no correlation between coach effect and performance, we calculated the expected frequency E for each cell in Table \ref{tab:5.2.1}.

Since the total sample with all expected frequencies is greater than 5, the conditions for using the chi-square test are met\cite{Cox2002}. Therefore, we used the chi-square test to assess the difference between the observed and expected frequencies. The calculated chi-square statistic $\chi^2 = 3.6000$, indicating that the observed frequencies are significantly different from the expected frequencies; the significance level $P = 0.0289$, indicating the rejection of the null hypothesis $H_0$ at the 95\% confidence level, \textbf{suggesting that there is a significant correlation between great coach effect and team performance.}

\subsection{The Impact of the Great Coach Effect}

To measure the impact of the “great coach effect,” we continued to select the U.S. women’s gymnastics team, where the coach effect is known to be significant, and analyzed the time period when they hired great coaches.

Using the ARIMA-LSTM model constructed in Task 1, we can predict the number of medals obtained during the periods when each country hired a great coach. For the predicted values corresponding to the periods with a great coach, we can calculate the RMSE (Equation \ref{eq:rmse}) during the coaching period, denoted as $ RMSE_{coach} $, by comparing the predictions with the actual values. For the periods without a great coach, we can calculate the baseline $RMSE$, denoted as $ RMSE_{base} $. Subsequently, we can define the effect size of the great coach as:

\begin{equation}
    Effect = RMSE_{coach} - RMSE_{base}
\end{equation}

Here, $Effect$ represents the average deviation in medal predictions per competition caused by the coach factor, which reflects the impact of a great coach on the number of medals.

\begin{figure}[htp]
    \centering
    \includegraphics[width=0.6\linewidth]{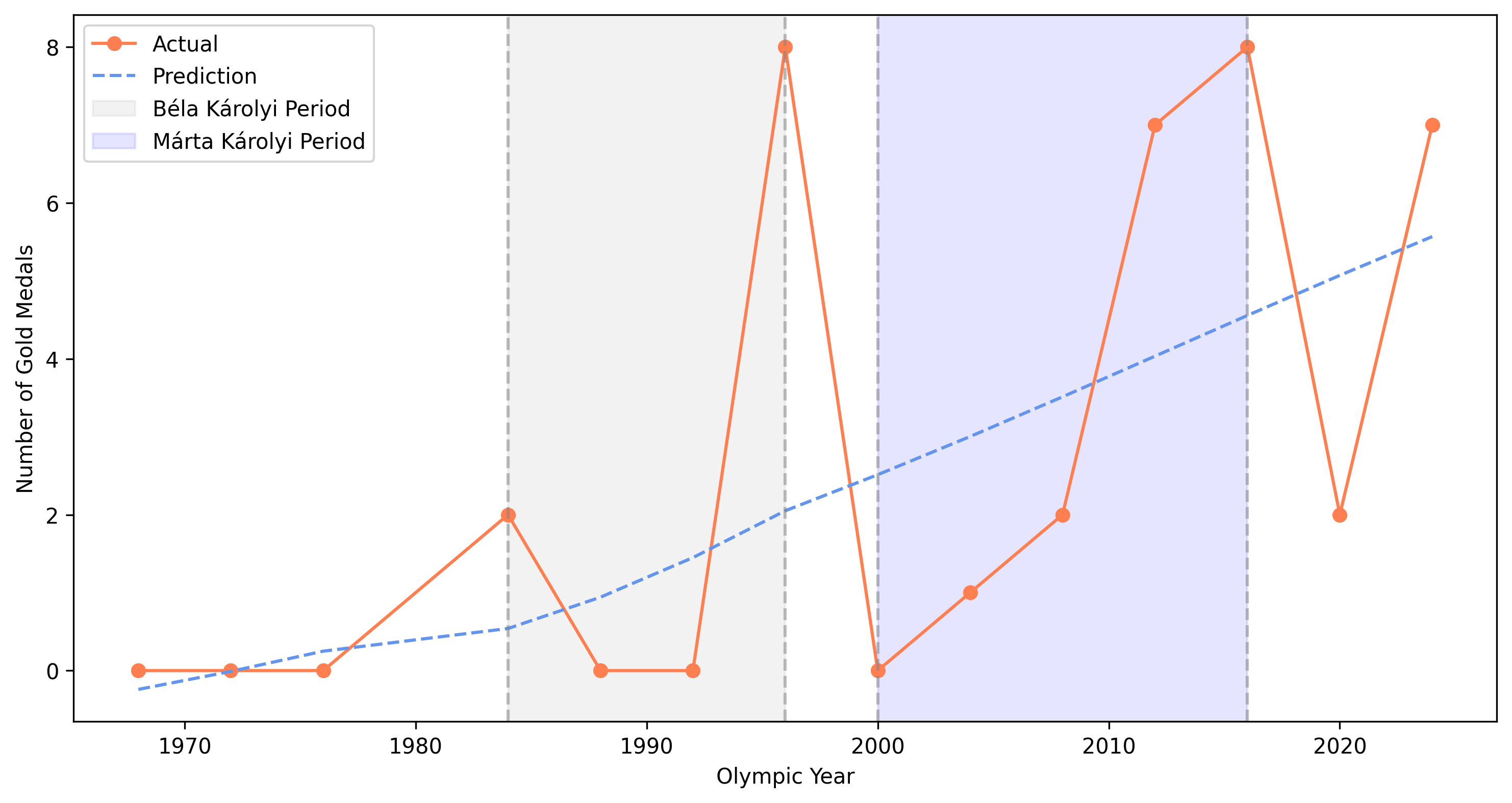}
    \caption{USA Women's Gymnastics Olympic Gold Medals: Actual vs Predicted}
    \label{fig:lstm}
\end{figure}

According to our model, we predicted the gold medals won by the U.S. women's gymnastics team and plotted Figure \ref{fig:lstm} to calculate the great coach effect $Effect$ by stage and true value. Among them, during the coaching period of the great coaches Béla Károlyi (1984-1996) and Márta Károlyi (2001-2016), the $RMSE_{coach}=2.86$ ; and during the period without the guidance of the great coach, the $RMSE_{base}=0.82$. Based on this, we calculated the Coach $Effect =2.04$, that is, \textbf{the great coach coaching period can win 2.04 more gold medals per Olympic Games on average than the period without the great coach coaching period.}

\subsection{Analysis of Great Coach Investment Strategy}

To explore the optimal investment strategy for hiring great coaches, we selected three representative countries—United States, France, and Italy—from the top 10 nations predicted to win the most medals at the 2028 Olympics. Based on the state matrix $M$ of national teams at specific time points, as proposed in Section \ref{subset:timestep_intro}, we defined the performance feature vector $N$ for each country across 71 Olympic sports. By calculating the $L2$ norm of the feature vector for each event, we obtained the performance feature value $V_p$ of the country in each event. The magnitude of the feature value reflects the overall performance level of the athletes from that country in a given event, where a higher feature value indicates a stronger competitive advantage in that event.

Based on the performance feature values, we further introduced the hypothesis of the "Great Coach Effect":

    \textbf{Assumption:} A large, sustained increase in medal totals over a short period, or a large, sustained decline over a period of time, can be attributed to the influence of great coaching.
    
To quantify this effect, we defined the "Coach Effect Coefficient" $E_{coach}$ as follows:

\begin{equation}
    E_{coach} = \max_{t} \left| \frac{M_t - AVG(M_{t-4:t-1})}{AVG(M_{t-4:t-1})} \times \frac{Count(M_t \diamond AVG(M_{t-4:t-1}))}{4} \right|
\end{equation}

where:

\begin{itemize}
    \item $M_t$: the number of medals won in year $t$;
    \item $AVG(M_{t-4:t-1})$: the average number of medals won from year $t-4$ to $t-1$;
    \item \(\diamond\): the comparison operator, defined as:
      \[
      \diamond = 
      \begin{cases} 
      >, & \text{if } \frac{M_t - AVG(M_{t-4:t-1})}{AVG(M_{t-4:t-1})} > 0 \\ 
      <, & \text{if } \frac{M_t - AVG(M_{t-4:t-1})}{AVG(M_{t-4:t-1})} < 0
      \end{cases}
      \]
\end{itemize}

Specifically, this coefficient is computed as the product of two critical components: (1) relative performance variation $\frac{M_t - AVG(M_{t-4:t-1})}{AVG(M_{t-4:t-1})}$  , measured as the percentage deviation from the mean of the preceding four years, and (2) a sustainability factor $\frac{Count(M_t \diamond AVG(M_{t-4:t-1}))}{4}$, represented by the proportion of subsequent years (up to four years) during which the performance maintains its directional trend. 

By simultaneously considering both immediate performance enhancement or decline and its temporal persistence, this methodology enables a comprehensive evaluation of the transformative impact of great coaches.

Based on the Coach Effect Coefficient, we proposed a balanced metric to evaluate the need for hiring great coaches, referred to as the Coach Impact Index $Index_{coach}$, which is expressed as:

\begin{equation}
    Index_{coach} = V_p \times (1 + E_{coach})
\end{equation}

The higher the $Index_{coach}$, the stronger the team's ability base is and the more sensitive it is to the great coach effect, so it is more necessary to hire an great coach to further improve its performance in a specific event.

Based on the above formula, we calculated the Coach Impact Index $Index_{coach}$ for the United States, France, and Italy. Subsequently, we ranked the top three sports disciplines in each country that require investment in great coaches according to the magnitude of the index, as shown in the Table \ref{tab:5.3}:

\begin{table}[htpb]
\centering
\renewcommand{\arraystretch}{1.2}
\caption{Top three sports disciplines requiring investment in great coaches}
\label{tab:5.3}
\begin{tabular}{cccc}
\toprule
  & \textbf{1st Program}  & \textbf{2nd Program}  & \textbf{3rd Program} \\ \midrule
\textbf{United States} & Athletics & Swimming & Artistic Gymnastics \\
\textbf{France} & Fencing & Judo & Handball \\ 
\textbf{Italy} & Fencing & Cycling Track & Shooting \\ 
\bottomrule
\end{tabular}
\end{table}

After sorting, the United States shows great investment potential in \textbf{Athletics, Swimming and Artistic Gymnastics}; France focuses on \textbf{Fencing, Judo and Handball}; Italy focuses on \textbf{Fencing, Cycling Track and Shooting}.

By investing in great coaches for these projects, \textbf{it is expected that the number of medals can be significantly improved in the short term}, as the national teams have a solid foundation in these areas and are highly sensitive to the effect of great coaches.

\section{Task3: Interesting Findings}  % 一级标题

\subsection{Analysis of participation trends of male and female athletes}

In data analysis, our model can clearly understand the historical gender distribution of Olympic medal winners.

\begin{figure}[htbp]
    \centering
    \begin{minipage}[b]{0.45\textwidth}
        \centering
        \includegraphics[width=\textwidth]{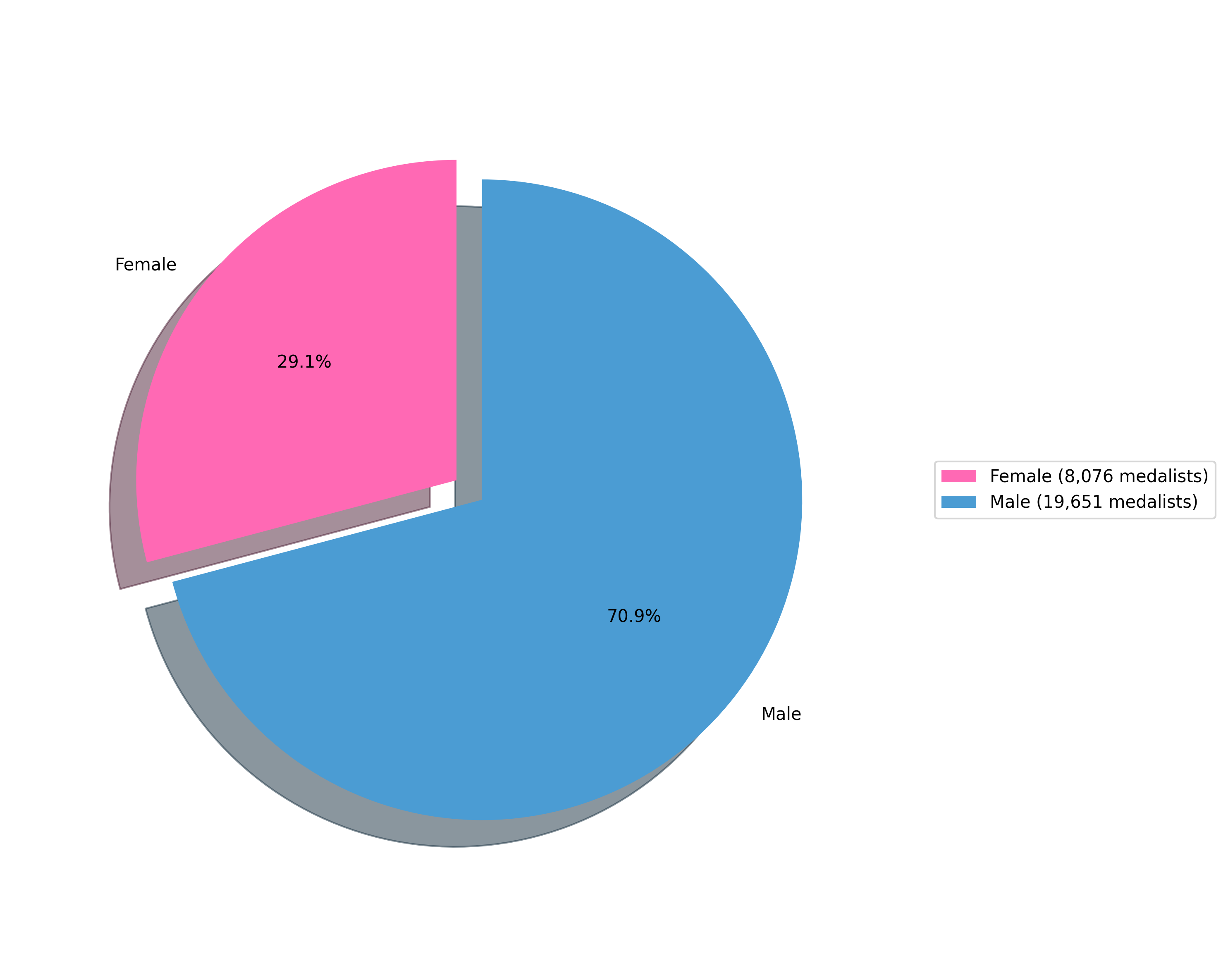}
        \caption{Gender medalists distribution pie chart}
        \label{fig:gender_analysis_pie}
    \end{minipage}
    \hfill
    \begin{minipage}[b]{0.45\textwidth}
        \centering
        \includegraphics[width=\textwidth]{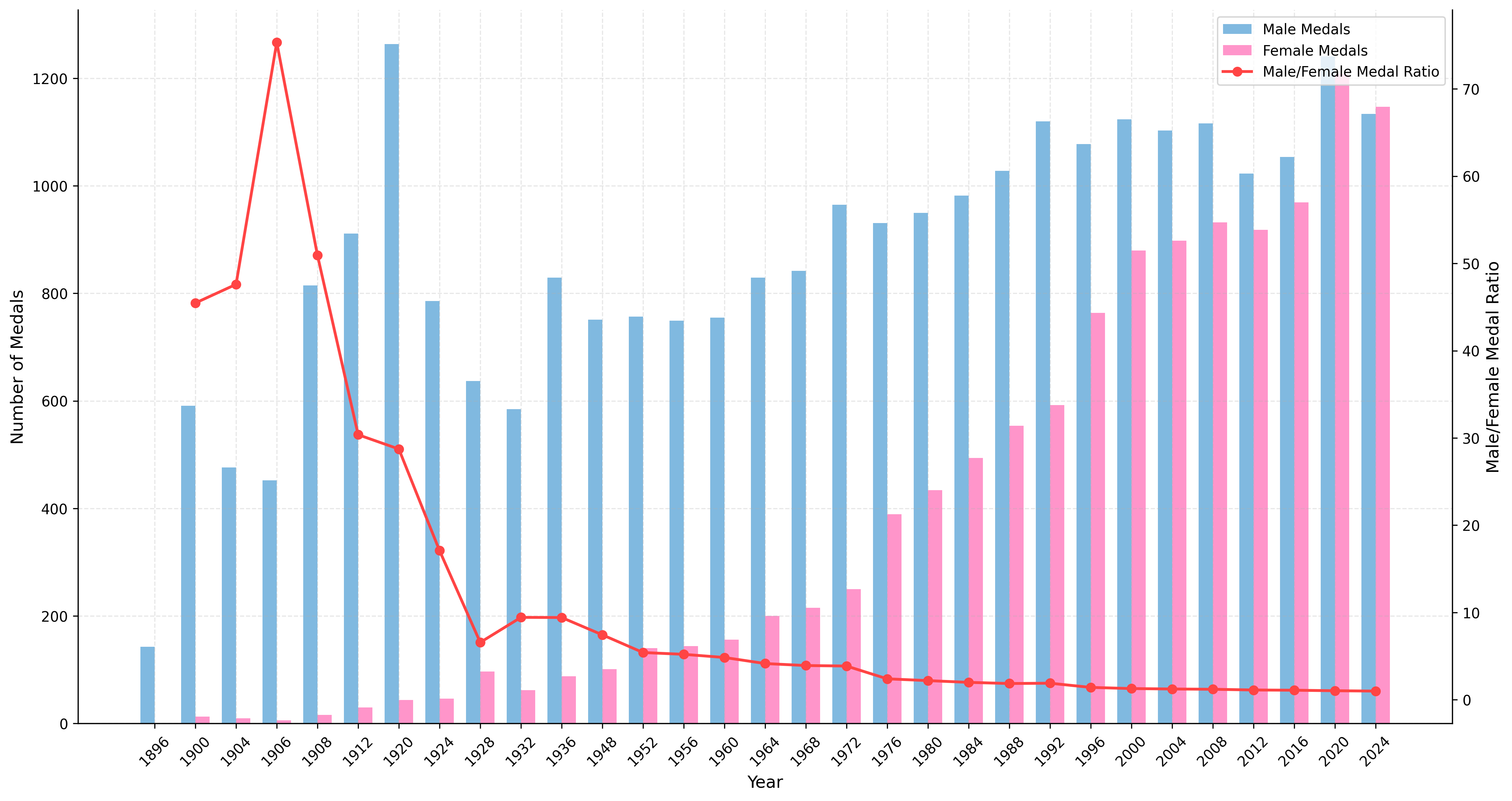}
        \caption{Medals by gender}
        \label{fig:gender_analysis}
    \end{minipage}
    \caption{Olympic Medals Gender Analysis}
    \label{fig:gender_combined}
\end{figure}

From Figure \ref{fig:gender_combined}, the trend analysis from 1896 to 2016 demonstrates a significant decline in the male-to-female ratio of Olympic medalists (indicated by the red line), decreasing from approximately 70:1 in the early years to nearly 1:1 in recent times, reflecting substantial progress in gender equality at the Olympic Games. The aggregate data reveals that male athletes (19,651) and female athletes (8,076) collectively represent a substantial scale of Olympic achievement, with males accounting for 70.9\% and females 29.1\% of total medalists.

\textbf{Recommendation}: There remains considerable potential for development in women's sporting events. Investment in women's athletics not only promotes gender equality but also presents opportunities for nations to achieve competitive advantages in emerging sporting disciplines. Therefore, enhancing support and development programs for female athletes should be considered a strategic priority for improving Olympic competitiveness.

\subsection{The impact of "traditional advantages"}

Traditional advantage can be defined as the potential influence factors that a country has in a specific sport. This influence factor is difficult to quantify through explicit indicators, but it exists objectively. This advantage is often reflected in the sustained competitive advantage of certain national teams in specific events, such as long-term leading positions in gymnastics and other events. After considering the number of participants, host advantage and historical results, our model uses the SHAP (SHapley Additive exPlanations) method to quantify the marginal contribution of each team's traditional advantages to the predicted results.

To measure the impact of traditional advantages on the predicted value, we have the formula

\begin{equation}
    \phi_{\text{team}}(\boldsymbol{x})=\sum_{S\subseteq F\setminus\{\text{team}\}}\frac{|S|!(|F|-|S|-1)!}{|F|!}[f_{S\cup\{\text{team}\}}(\boldsymbol{x})-f_S(\boldsymbol{x})]
\end{equation}

where $F$ denotes the set of all feature variables (including number of participants, home advantage, historical performance, etc.), S represents the feature subset excluding team traditional advantages, $f_{S\cup\{\text{team}\}}$ and $f_S$ represent the predicted results with and without team traditional advantages, respectively. By calculating the weighted average of the marginal contributions of all possible subset combinations, we get the Shapley value of the team traditional advantage. Then we list the combinations of projects where the traditional advantage has a greater impact on the final results, and we can get the traditional advantage projects of each country. Table \ref{tab:traditional_adv} shows the combinations of teams and projects where the team traditional advantage exceeds 0.3 in explaining the gold medal prediction.

\begin{table}[htpb]
\centering
\renewcommand{\arraystretch}{1.2}
\caption{Projects with Significant Team Traditional Advantages in Gold Medal Predictions}
\label{tab:traditional_adv}
\begin{tabular}{ccc}
\toprule
\textbf{Program} & \textbf{Country} & \textbf{Shapley value} \\ \midrule
Taekwondo & South Korea & 0.6221 \\
Wrestling & Russia & 0.5845 \\
Diving & China & 0.5279 \\
Fencing & Russia & 0.4865 \\
Tennis & United States & 0.3741 \\
\bottomrule
\end{tabular}
\end{table}

\textbf{Recommendation}: National Olympic Committees should adopt the strategy of "focusing on strengths and balanced development": while maintaining continuous investment in traditional advantage projects (such as Korean taekwondo and Chinese diving), they should also pay attention to the development potential of emerging projects. Specifically, for traditional advantage projects with high Shapley values (>0.5), resources should be maintained or increased to consolidate competitive advantages; for projects with medium Shapley values (0.3-0.5), it is recommended to evaluate the room for improvement and invest in a targeted manner to maximize the medal winning rate.

\section{Sensitivity Analysis}  % 一级标题

In order to verify the rationality of the ARIMA-LSTM model and its dependence on training data, we randomly selected some data for training and analyzed the impact of changes in data volume on the stability and accuracy of the prediction results. The training samples in this article come from the relevant data sets of the Olympic Games from 1896 to 2024, which contain awards from about 233 countries and regions and more than 250,000 athlete participation records. Since the model mainly relies on the country's historical awards and athlete participation information as input, we extracted 75\% and 50\% of the athlete participation data and 75\% and 50\% of the historical year award records from the original sample respectively, and combined them with the original data set Different combinations formed 9 reduced data sets, and the prediction accuracy was tested respectively.

\begin{figure}[htp]
    \centering
    \includegraphics[width=0.6\linewidth]{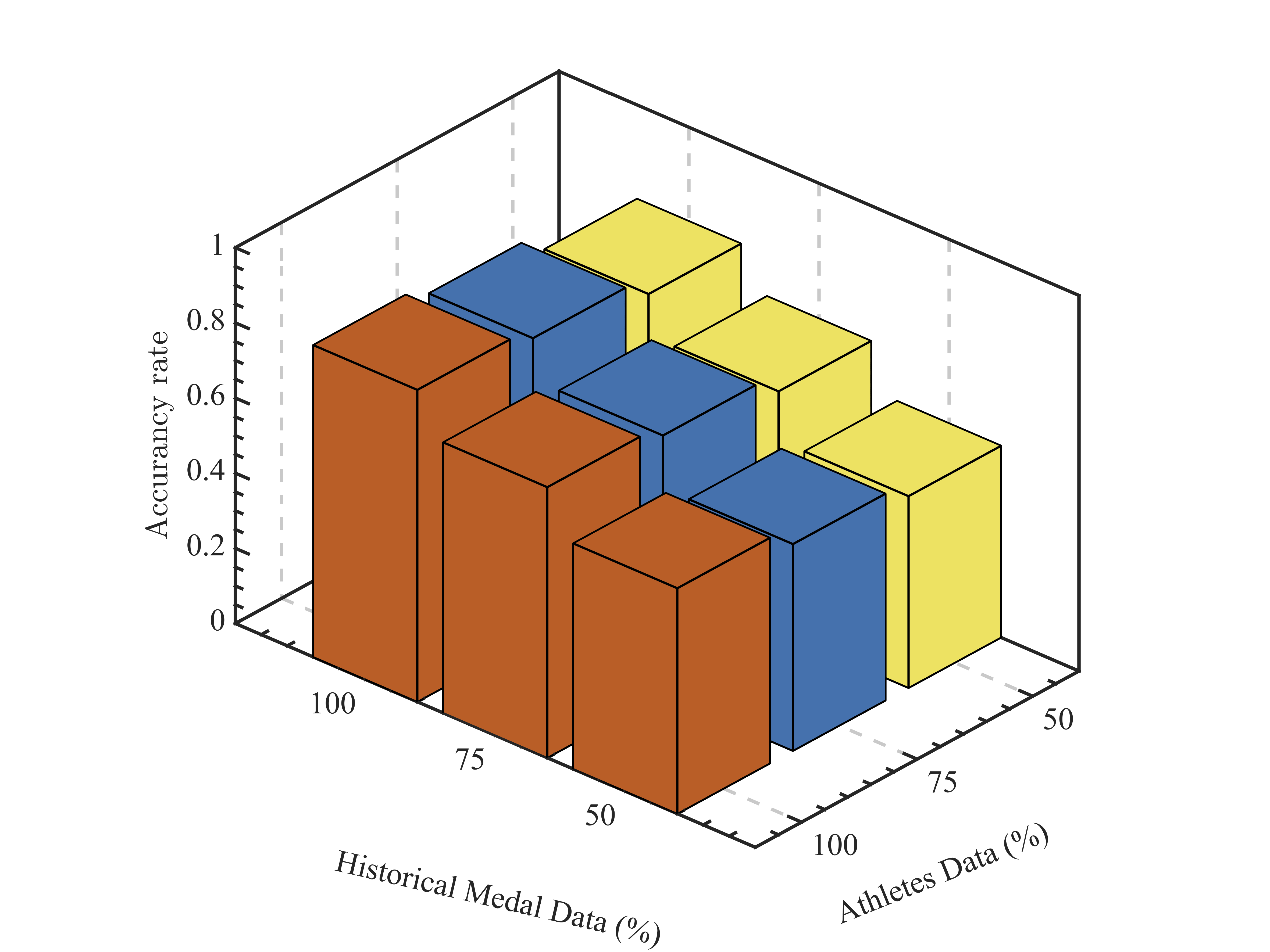}
    \caption{Classification Accuracy under Different Data Coverage}
    \label{fig:sensitive}
\end{figure}

As can be seen from Figure \ref{fig:sensitive}, since the LSTM model is more dependent on the amount of training data, it shows good classification results on the data set with the largest amount of data, with an accuracy of 83\%. As the athlete participation and historical data coverage decrease, the classification accuracy continues to decrease, especially in the case of 50\% athletes and 50\% historical data, the accuracy is only 51\%. In various combinations with a data coverage of around 75\%, the accuracy fluctuates slightly between 64\% and 72\%, indicating that our model has good robustness and is suitable for predictions with higher data coverage (above 75\%).

% \printbibliography  % 打印引用文献列表


%%%%%%%%%%%%%%%%%%%%%%% 正文结束 %%%%%%%%%%%%%%%%%%%%%%%

% \begin{appendices}  % 附录

% \section{appendix}  % 一级标题

% \textbf{Python source code:}
% \lstinputlisting[language=C++]{./code/python.py}

% \end{appendices}  % 附录结束
\end{document}